\documentclass[preprint,12pt]{elsarticle}

\usepackage{amssymb}
\usepackage{subcaption}

\journal{ }

\begin{document}

\begin{frontmatter}



\title{Improving correlation method with convolutional neural networks}


\author[label1]{Dmitriy S. Goncharov}
\ead{goncharov.dms@yandex.ru}
\author[label1]{Rostislav S. Starikov}

\address[label1]{National Research Nuclear University "MEPhI"}

\begin{abstract}
We present a convolutional neural network for the classification of correlation responses obtained by correlation filters. The proposed approach can improve the accuracy of classification, as well as achieve invariance to the image classes and parameters.
\end{abstract}




\end{frontmatter}


\section{Introduction}

Today, various computer vision technologies are increasingly used to solve problems such as object detection and classification. Convolutional neural networks, which are powerful and high-precision tools that allows you to solve complex problems, are used in most cases. However, there are simpler methods that work well when solving simpler problems, for example, the correlation method.

The invariant correlation method is a simple method that allows you to perform binary classification of objects, as well as detect their location \cite{VijayaKumar2005}. The simplicity of the correlation method allows one to achieve high processing speed. However the correlation method doesn't work well when solving complex tasks such as multiclass image classification.

The main idea of this work is to use a convolutional neural network to analyze the shape of the received correlation responses. We hope, that the combination of high speed of correlation matching and flexibility of neural networks is of interest.

\section{Method description}

The correlation method works as follows. A grayscale image or any channel of a color image can be represented as a two-dimensional signal $f(x, y)$. By calculating the cross correlation of signal $f(x, y)$ and some reference object $h(x, y)$, you can get a correlation response, by the size and shape of which you can classify the object as true or false, and also you can determine the location of the object of interest by the correlation peak location.

Using the convolution theorem, the cross correlation of signals $f(x, y)$ and $h(x, y)$ can be represented as follows:

\begin{equation}\label{eq_cor_ft}
	f\star h = \mathcal{F}^{-1}(F^*\cdot H),
\end{equation}
\noindent
where $\mathcal{F}^{-1}$ is the operator of the two-dimensional inverse Fourier transform, $F$ and $H$ are the signals $f$ and $h$ in the Fourier domain respectively. This representation can significantly accelerate the calculation of the correlation between large matrices due to the use of the fast Fourier transform. Also this representation allows us to fully implement the correlation method in optical system with very high speed \cite{VijayaKumar2005}.

\begin{figure}[ht!]
    \centering
    \includegraphics[scale=0.6]{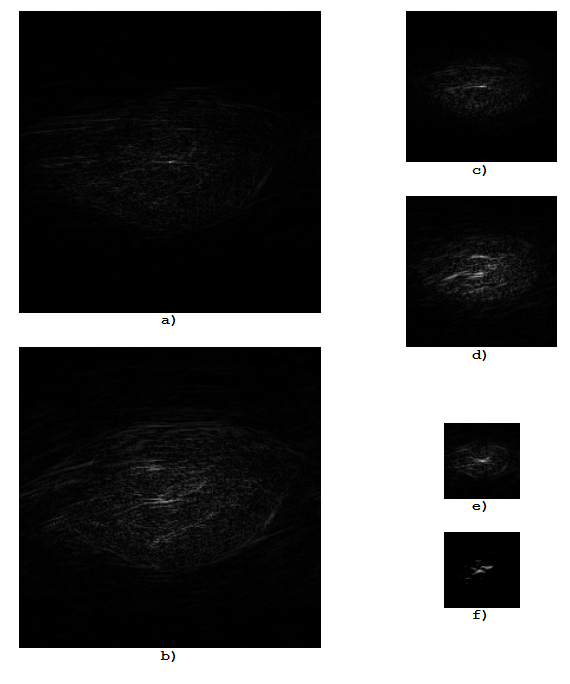}
    \caption{Correlation responses: (a) --- true class, $256\times 256$; (b) --- false class, $256\times 256$; (c) --- true class, $128\times 128$; (d) --- false class, $128\times 128$; (e) --- true class,  $64\times 64$; (f) --- false class,  $64\times 64$.}
    \label{fig_CORR_PEAKS}
\end{figure}

Applying of invariant correlation filters allows you to provide a recognition at various changes of the input object (change in angle, scale, illumination, etc.) \cite{VijayaKumar2005}. In this work we used two different types of correlation filters: OT MACH (optimal tradeoff maximum average correlation height) \cite{Mahalanobis1994} and MINACE (minimum average correlation energy) \cite{Ravichandran1992} filters. These filters are synthesized based on training data and optimize different parameters, but allow achieving almost the similar shape of the correlation peak which is close to the delta function for true objects. Figure \ref{fig_CORR_PEAKS} shows examples of correlation responses obtained for images of various objects in various resolutions.


There are some  simple metrics that are usually used to analyze correlation responses, for example maximum height of the correlation peak and PCE (peak to correlation energy) \cite{VijayaKumar2005}. However, these metrics analyze to a greater extent only the value of the correlation response, while its shape is almost not taken into account. Unlike using the metrics mentioned above, we offer to use the convolutional neural network (CNN) to analyze only the shape of correlation peaks.

\section{Convolutional neural network}

\subsection{Data preprocessing}

The CNN must accurately classify any correlation responses. Since we use correlation filters that provide for true objects a correlation response of a shape close to the delta function, it was decided to use only the central part of the correlation responses of size 32 x 32. We assumed, that this approach ensures invariance for input image resolution. Also each correlation response was normalized so that all the data was distributed from 0 to 1.

\subsection{Architecture}

In this work, we used a modified version of the CNN ResNet-18 \cite{HeZhangSun2016, he2016identity}. The number of trained parameters was reduced to 1.2 million by reducing the number of filters in all convolutions. The maxpooling layer was removed. Input images are 32 x 32 images in one color channel. The last layer is a fully connected layer with 1 neuron with sigmoid activation at the output. 

We experimented with various ReLU-like activations \cite{NairV.2010}, and in this work we settled on Swish activation \cite{ramachandran2017searching}. Also we widely used Batch Normalization in this CNN \cite{Ioffe2015}. The first layer of our CNN is Batch Normalization layer, which allows us to feed already normalized data to the input of the first convolutional layer.

The CNN architecture was implemented in python 3 using Keras and TensorFlow frameworks.

\subsection{Training}

\begin{figure}[ht!]
	\centering
	\includegraphics[scale=0.4]{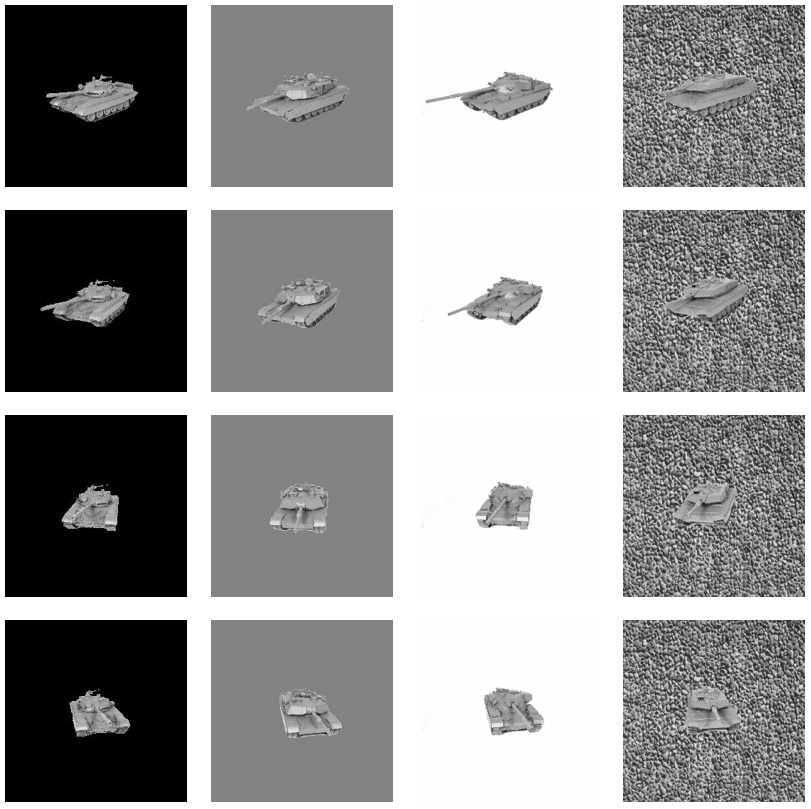}
	\caption{Examples of images of tanks.}
	\label{fig_TANKS}
\end{figure}

Correlation responses obtained on the images of tanks (Figure \ref{fig_TANKS}) using OT MACH CF were used as the data for CNN training. In total, this set contains 2880 images of the T-72 tank (true object) and 1440 images of Abrams, Leopard and Chieftain tanks (false objects) each. Images are presented in different resolutions (from 256 x 256 to 32 x 32), the tanks are located on a different backgrounds and rotated at different angles. As training data for the CNN, correlation responses corresponding to the T72, Abrams, and Leopard tanks were selected (4320 images in total). As test data, correlation responses corresponding to the T72 and Chieftain tanks (2880 images) were selected.

To prevent overfitting, various augmentations were used: flips along the horizontal and vertical axis, rotation by a random angle from 0 to 90 degrees, adding normal Gaussian noise with a random standard deviation. Also the regularization of the L2-norm with a coefficient of 0.005 was used.

Over 10 epochs of training, it was possible to achieve a classification accuracy of test images of 92 \%. At the same time, the classification accuracy is highly dependent on image resolution. Images with higher resolution are classified with 100\% accuracy, while images with the lowest resolution are classified with an accuracy of about 80\%.

\section{Experiments}

For all images from the set above (Figure \ref{fig_TANKS}), new correlation responses were obtained using the MINACE CF. Table \ref{tab1} shows the results obtained by processing all correlation responses obtained by both OT MACH and MINACE CFs using peak height and PCE metrics and CNN. Total number of used correlation responses is 14400. It is easy to see that the CNN provides much higher accuracy than standard correlation metrics.

\begin{figure}[ht!]
	\centering
	\includegraphics[scale=0.4]{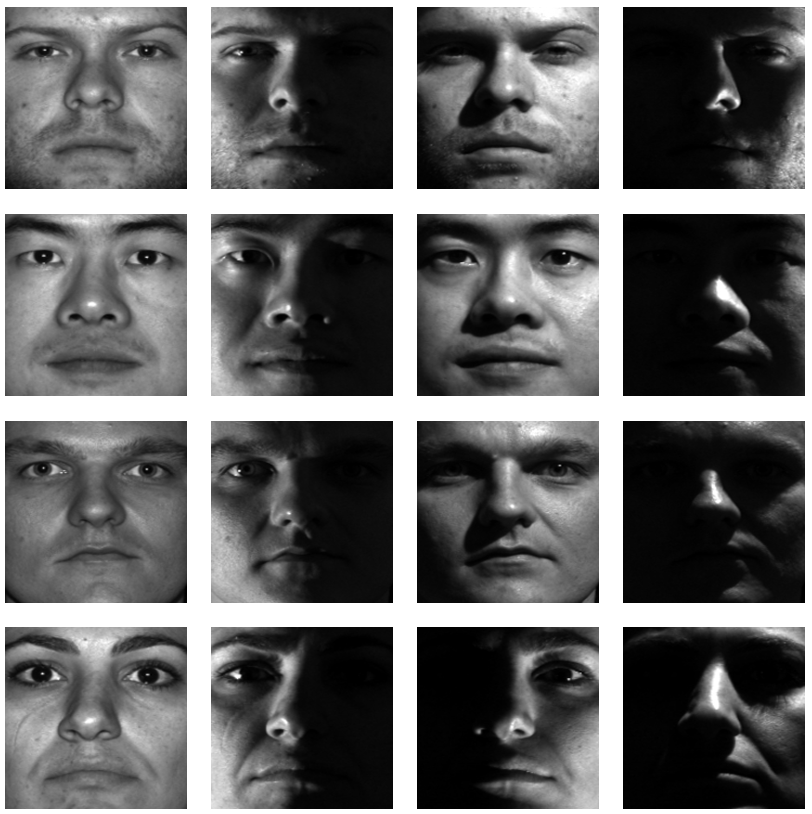}
	\caption{Sample face images from the Yale Face Database B.}
	\label{fig_FACES}
\end{figure}

It is also very interesting to check if a trained CNN can classify correlation responses received from images of completely different objects. For this, the Yale Face Database B \cite{GeBeKr01} dataset was used, examples of images of which are shown in Figure \ref{fig_FACES}. 3400 new correlation responses were obtained on the images from this dataset using both OT MACH and MINACE CFs. Table \ref{tab2} shows the results obtained by processing these correlation responses. As we can see, in this case, the CNN shows the highest efficiency, despite the fact that it was trained on the correlation responses obtained from images of a different type. Or shortly: the CNN, trained on tanks responses, provides accurately classification of faces responses.

\begin{table}[ht!]
    \caption{Results of processing correlation responses from tanks images by different methods.}
    \label{tab1}
    \resizebox{\textwidth}{!}{%
    \begin{tabular}{|c|c|c|c|}
    \hline
    \begin{tabular}[c]{@{}c@{}}Post processing\\  method\end{tabular}                               & \begin{tabular}[c]{@{}c@{}}Correlation peak\\  height\end{tabular} & \begin{tabular}[c]{@{}c@{}}PCE\\ (peak to correlation\\ energy)\end{tabular} & ResNet-18 \\ \hline
    \begin{tabular}[c]{@{}c@{}}Number of image sets \\ with an error \textless 0,001\%\end{tabular} & 8                                                                  & 8                                                                            & 19        \\ \hline
    \begin{tabular}[c]{@{}c@{}}Number of image sets \\ with an error \textgreater 25\%\end{tabular} & 7                                                                  & 6                                                                            & 0         \\ \hline
    \begin{tabular}[c]{@{}c@{}}Average error of \\ the other sets, \%\end{tabular}                  & 7.42                                                               & 5.15                                                                         & 7.95      \\ \hline
    \end{tabular}%
    }
\end{table}

\begin{table}[ht!]
    \caption{Results of processing correlation responses from faces images by different methods.}
    \label{tab2}
    \resizebox{\textwidth}{!}{%
    \begin{tabular}{|c|c|c|c|}
    \hline
    \begin{tabular}[c]{@{}c@{}}Post processing\\  method\end{tabular} & \begin{tabular}[c]{@{}c@{}}Correlation peak\\  height\end{tabular} & \begin{tabular}[c]{@{}c@{}}PCE\\ (peak to correlation\\ energy)\end{tabular} & ResNet-18 \\ \hline
    Average error, \%                                                 & 10.52                                                              & 7.67                                                                         & 5.41      \\ \hline
    \end{tabular}%
    }
\end{table}

\newpage
\section{Conclusion}

As we see, training a neural network to classify correlation responses obtained by invariant correlation filters allows to achieve the following: invariance to image resolution, invariance to the type of recognizable images, some independence to the correlation filter type used (with a given shape of the correlation responses) and high accuracy of correlation responses classifying. 




\bibliographystyle{elsarticle-num} 
\bibliography{biblio}




\end{document}